\documentclass[twoside,11pt]{article}

\usepackage{blindtext}

\usepackage{jmlr2e}
\usepackage{bm}
\usepackage{amsmath}
\usepackage{subfigure}
\usepackage{booktabs}

\usepackage{lastpage}

\ShortHeadings{Fractal Flow}{Binhui Zhang and Jianwei Ma}

\firstpageno{1}

\begin{document}

\title{Fractal Flow: Hierarchical and Interpretable Normalizing Flow via Topic Modeling and Recursive Strategy}

\author{\name Binhui Zhang \email 25B312019@stu.hit.edu.cn \\
	\addr School of Mathematics\\
	Harbin Institute of Technology, Harbin 150001, China
	\AND
	\name Jianwei Ma \email jwm@pku.edu.cn \\
	\addr School of Mathematics and Institute for Artificial Intelligence\\
	Harbin Instiute of Technology, Harbin 150001, China\\
	School of Earth and Space Sciences and Institute for Artificial Intelligence\\
	Peking University, Beijing 100871, China}

\editor{My editor}
\maketitle

\begin{abstract}
Normalizing Flows provide a principled framework for high-dimensional density estimation and generative modeling by constructing invertible transformations with tractable Jacobian determinants. We propose Fractal Flow, a novel normalizing flow architecture that enhances both expressiveness and interpretability through two key innovations. First, we integrate Kolmogorov–Arnold Networks and incorporate Latent Dirichlet Allocation into normalizing flows to construct a structured, interpretable latent space and model hierarchical semantic clusters. Second, inspired by Fractal Generative Models, we introduce a recursive modular design into normalizing flows to improve transformation interpretability and estimation accuracy. Experiments on MNIST, FashionMNIST, CIFAR-10, and geophysical data demonstrate that the Fractal Flow achieves latent clustering, controllable generation, and superior estimation accuracy.
\end{abstract}

\section{Introduction}

Finding an accurate probabilistic model---one that can discover the high-level latent structure of data and correctly describe the data generation process---has long been a fundamental goal in both statistical science and machine learning. In recent years, deep generative models have emerged as powerful tools for probabilistic modeling, capable of capturing complex, high-dimensional data distributions. Among various deep generative paradigms, diffusion models \citep{ho2020denoising,rombach2022high} have gained significant popularity in recent years due to their strong generative capabilities. However, iterative denoising steps in diffusion models make both training and inference computationally intensive, and these models do not provide tractable or explicit probability density estimation. Another class of models, autoregressive generative models \citep{van2016pixel,van2016conditional,parmar2018image} offer an alternative by modeling the discrete distribution of quantized image pixels directly. This approach avoids the need for iterative refinement and enables fast, parallelizable training with exact likelihood computation. However, since both inference and generation in autoregressive models are inherently sequential, the overall speed remains a significant limitation.

Flow-based generative models, especially those built on Normalizing Flows (NFs) \citep{dinh2017realnvp,kingma2018glow,papamakarios2017masked,durkan2019neural}, not only enable exact probability density estimation, but also support efficient sampling. These models transform a simple base distribution into complex data distributions through a sequence of invertible and differentiable mappings. However, the requirement of invertibility and tractable Jacobian determinants imposes structural constraints on model design, often limiting their expressiveness---particularly for high-dimensional data like natural images. As a result, flow-based models had long remained less popular than diffusion and autoregressive approaches in image generation tasks. Recent studies on flow-based generative models, such as MeanFlow \citep{geng2025mean} and STARFlow \citep{gu2025starflow}, have significantly expanded the speed advantages with sample quality approaching that of state-of-the-art diffusion models. As a result, flow-based models are once again gaining attention as competitive alternatives in the field of deep generative modeling.

In the field of flow-based models, coupling-based normalizing flows hold a significant position due to their tractable Jacobian determinants, efficient computation speed, and scalable architecture. Current research on coupling-based flows mainly focuses on two directions: designing more expressive and efficient invertible transformations \citep{kolesnikov2024jet} and improving the structure and functionality of the latent space. In the field of Variational Autoencoders (VAEs), a variety of works such as $\beta$-VAE \citep{higgins2017beta} and TCVAE \citep{kumar2018variational} have extensively explored how to promote disentangled latent representations. These studies consistently demonstrate that disentanglement can lead to several desirable properties, including improved interpretability and better transferability. For NFs, a disentangled latent space not only brings these benefits, but also facilitates the learning of the invertible transformation $\mathbf{z} = T(\mathbf{x})$, making it more interpretable. Although a few works such as \cite{izmailov2020semi} and \cite{yao2024hierarchical} have incorporated Gaussian mixture priors or hierarchical latent structures into normalizing flow models, these explorations are primarily task-specific (e.g., semi supervised learning or anomaly detection). A general, learnable, and interpretable framework for latent space construction in NFs remains underdeveloped.

Motivated by these limitations, we turn our focus to the construction of latent spaces in NFs, aiming to develop a general and learnable architecture that enhances both expressiveness and interpretability. Recently, Kolmogorov–Arnold Networks (KANs) \citep{liu2024kan} have gained increasing attention as an interpretable alternative to Multilayer Perceptrons (MLPs), offering both high flexibility and strong interpretability through learnable activation functions. We observe that with appropriate constraints, such learnable nonlinear layers are well-suited for modeling probability density functions. Through our study of the Kolmogorov–Arnold representation theorem (KAT), we observe a natural synergy between KANs and NFs. This integration expands the flexibility and interpretability of the latent space in NFs, enabling more structured representations. In such structured latent spaces, we can also leverage statistical modeling approaches, such as topic modeling, to further enhance structure and interpretability. Topic modeling is one of the most powerful techniques in text mining for data mining, latent data discovery, and finding relationships among data and text documents. It is often applied in natural language processing to topic discovery and semantic mining from unordered documents \citep{chauhan2021topic}. Latent Dirichlet Allocation (LDA) \citep{blei2003latent} is particularly well-suited for modeling the semantic structure of latent spaces, as it provides a principled statistical framework for capturing hierarchical topics. Additionally, we find inspiration from Fractal Generative Models (FGMs) \citep{li2025fractal}, which abstract a generative model itself as a module and recursively invokes generative modules of the same kind within itself to construct complex architectures. This recursive strategy can be effectively adapted to the invertible transformations inNFs, enhancing both model interpretability and estimation accuracy.

Inspired by the above observations, we propose Fractal Flow, a novel NF architecture that enhances both expressiveness and interpretability through two key extensions. First, we integrate the KAT with NFs by designing KAN-inspired layers to model the latent space. Building on this integration, we further incorporate LDA into the latent space as a statistical topic modeling approach, enabling the model to better capture hierarchical semantic topics. Second, we introduce a recursive strategy into NFs, which naturally forms a fractal structure, to further enhance the interpretability of the model. Unlike FGMs, which recursively invoke entire generative models as modules, our approach regards each invertible transformation from the original space to its latent space as a recursive module. Specifically, we model the latent space generated by one invertible transformation with another smaller-scale normalizing flow, and recursively repeat this process to form a hierarchy of latent spaces and transformations, until the final latent space can be effectively modeled with KAN-inspired layers and LDA. Empirical evaluations on MNIST and FashionMNIST demonstrate that Fractal Flow improves latent space clustering and interpretability, while achieving competitive or superior estimation accuracy under comparable parameter budgets. Additionally, our experiments systematically investigate the impact of different subnetwork choices, highlighting the advantages of convolutional structures. Visualization results further demonstrate that  Fractal Flow enables controllable generation and provides interpretability of the generation process, and we further validate its effectiveness on the automobile class of CIFAR-10 as well as geophysical seismic data \citep{ROG2021}, demonstrating its applicability to more complex and domain-specific scenarios.

\section{Method}

\subsection{Structured Latent Priors in Normalizing Flows}
In the context of NFs, the latent space is typically assumed to follow a simple, tractable prior distribution, such as a standard Gaussian  distribution. While this assumption facilitates model training, it often fails to capture the complex, hierarchical, or clustered structures inherent in real-world data, where distributions are often scattered into different clusters. When these clusters are isolated, the optimal invertible transformation $T(\mathbf{x})$ would be complex \citep{dong2022normalizingflowvariationallatent}, making it difficult to learn. As a result, there is a growing interest in exploring more expressive and structured latent representations that can better reflect the underlying data characteristics and enhance model interpretability. In VAEs, the latent space is a compressed, interpretable feature space that enables tasks such as controllable generation, clustering, and downstream transfer learning. This observation naturally leads to the question: why not apply a similar philosophy in NFs? Despite mapping input data into latent representations via invertible transformations, the latent spaces in most flow models remain unstructured and uninterpretable.

Traditional MLPs construct functions through a composition of linear transformations with learnable weights on edges and fixed nonlinear activations on nodes. However, this structure often limits their ability to form interpretable and flexible latent spaces, especially when modeling complex probability distributions. In contrast, KANs, as a powerful and interpretable alternative to MLPs, adopt a fundamentally different design: they rely on learnable activation functions on edges followed by sum operation on nodes. This architecture naturally supports the construction of learnable probability spaces. Furthermore, by replacing the original cubic spline in KANs with Gaussian radial basis functions, the resulting representation becomes directly interpretable in terms of Gaussian mixture models. This modification enhances the statistical meaning of the learned latent representations and facilitates probabilistic interpretation.

KANs are inspired by the KAT, which not only justifies the design of KANs, but also offers a principled perspective on structured function approximation in high dimensions. The KAT provides a foundational tool for expressing multivariate continuous functions through compositions of univariate functions \citep{kolmogorov1961representation, liu2024kan}. For a smooth $f : [0,1]^n \to \mathbb{R}$,
\begin{equation}
	f(x) = \sum_{q=1}^{2n+1} \Phi_q\left( \sum_{p=1}^n \phi_{q,p}(x_p) \right),
\end{equation}
where $\phi_{q,p} : [0,1]^n \to \mathbb{R}$ and $\Phi_q : \mathbb{R} \to \mathbb{R}$. Inspired by the KAT, we interpret the outer summation of functions as a form of latent mixture structure, analogous to classical probabilistic models such as Gaussian Mixture Models or latent variable formulations. In this view, each function can be associated with a specific mode or cluster in the latent space, providing a natural connection to hierarchical or structured prior distributions. The inner summation of functions in the classical KAT consists of univariate transformations applied independently across dimensions. However, in our approach, these inner functions are generalized to deep, invertible flow-based transformations, allowing for rich dependencies and interactions among variables while preserving tractability. When these outer components are instantiated with Gaussian functions, the inner functions operate independently across dimensions, mirroring the conditional independence assumptions commonly adopted.

Specifically, following the above interpretation, we reformulate the KAT into a structured latent prior Normalizing Flow model as follows:
\begin{equation}
	p(\mathbf{x}) = \sum_{k=1}^K w_k \, \mathcal{N}(\mathbf{z} \mid \mu_k, \Sigma_k) \left| \det \left( \frac{\partial \mathbf{z}}{\partial \mathbf{x}} \right) \right|,
\end{equation}
where $\mathcal{N}(\mathbf{z} \mid \mu_k, \Sigma_k)$ denotes the probability density function of a Gaussian distribution over the latent variable $\mathbf{z}$, with mean $\mu_k$ and covariance $\Sigma_k$, and $K$ is the number of Gaussian components in the mixture. Here, $\mathbf{z} = T(\mathbf{x})$ is a learnable, invertible transformation parameterized by a deep neural network, and constructed using a series of affine coupling layers as introduced in RealNVP. These transformations ensure invertibility and efficient computation of the Jacobian determinant, thereby enabling exact likelihood estimation and stable training.

To encourage structure-preserving transformations in flow, we introduce a transformation regularization term based on the Euclidean distance between $\mathbf{x}$ and $\mathbf{z}$. Unlike conventional flow models that only optimize likelihood, this regularization term penalizes excessive distortions of the data manifold, thereby preserving the structural consistency between the latent space and the data space.
\begin{equation}
	\mathcal{L}_{L_2} = \mathbb{E}_{\mathbf{x} \sim p(\mathrm{data})} \left[ \| \mathbf{z} - \mathbf{x} \|_2^2 \right].
\end{equation}
\begin{figure}[tbp]
	\centering
	\includegraphics[width=\textwidth]{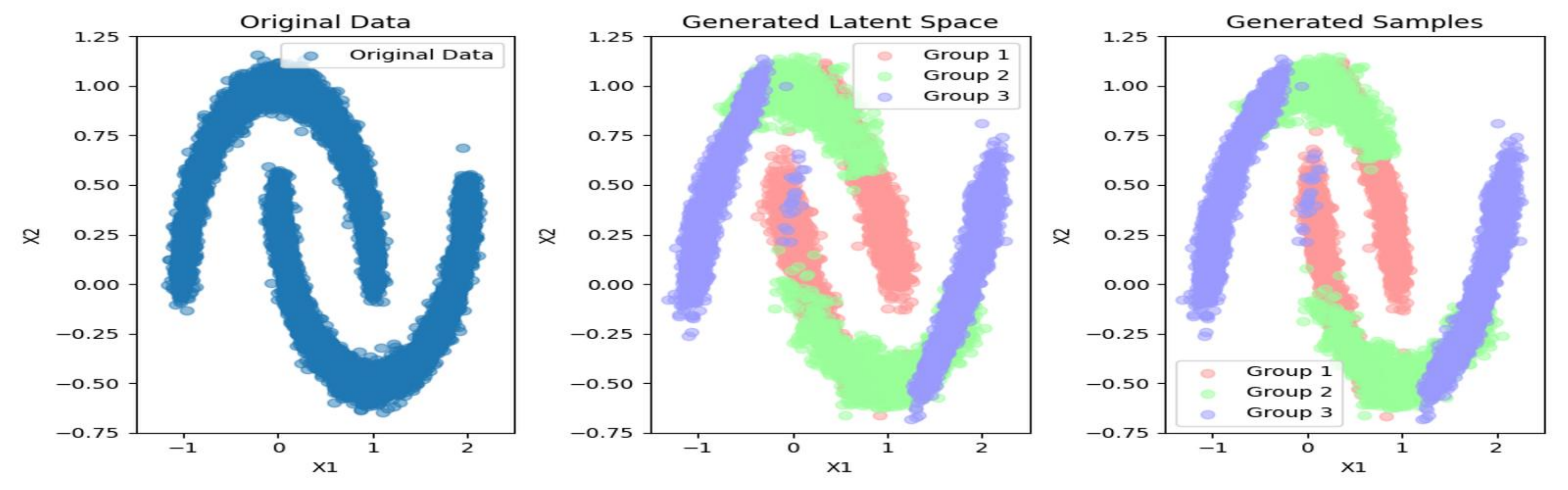}
	\caption{Visualization of a structured latent space using the Structured Latent Priors. The model effectively captures the underlying structure of the data.}
	\label{Figure:1}
\end{figure}

In high-dimensional settings, computing and learning full covariance matrices becomes computationally expensive and often intractable. However, the transformation $T(\mathbf{x})$, learned via deep invertible networks, is capable of capturing complex dependencies among variables. Therefore, we introduce an independence assumption in the latent space, treating the components of the latent variable z as statistically independent. This not only significantly reduces the complexity of the prior distribution but also aligns with the additive structure of inner functions in the KAT, which are applied independently across dimensions.
\begin{equation}
	p(\mathbf{x}) = \sum_{k=1}^K\frac{w_k \left| \det\left( \frac{\partial \mathbf{z}}{\partial \mathbf{x}} \right) \right|}{(2\pi)^{D/2} \, |\Sigma_k|^{1/2}} \exp\left\{ \sum_{d=1}^D -\frac{(z_d - \mu_{k,d})^2}{2\sigma_{k,d}^2} \right\}.
\end{equation}
In this formulation, each component in the structured normalizing flow corresponds directly to elements in the KAT. Specifically:
\begin{enumerate}
	\item $\sum_{k=1}^K \frac{w_k \left| \det\left( \frac{\partial \mathbf{z}}{\partial \mathbf{x}} \right) \right|}{(2\pi)^{D/2} \,|\Sigma_k|^{1/2}} \exp(\cdot)$ corresponds to the outer summation $\sum_{q=1}^{2n+1} \Phi_q(\cdot)$ in the KAT.
	\item $\sum_{d=1}^D -\frac{(z_d - \mu_{k,d})^2}{2\sigma_{k,d}^2}$  corresponds to the inner summation $\sum_{p=1}^n \phi_{q,p}(\cdot)$ in the KAT.
\end{enumerate}

\subsection{Neural Network Implementation}
KANs can be regarded as a neural network implementation of the KAT. In this framework, nonlinear activations are placed on edges, and simple summation is applied at the nodes, forming a direct correspondence to the KAT’s inner and outer function composition. The nonlinear activations used in KANs are implemented via grid-based B-spline interpolation, allowing each input dimension to be transformed flexibly using learnable piecewise functions. This grid-based structure enables KANs to capture fine-grained variations in each dimension while preserving interpretability and modularity.

Inspired by this design philosophy, we aim to retain the structure of nonlinear activations on edges and sum operation on nodes in our latent prior modeling. However, instead of using additive aggregation as in KANs, we adopt a multiplicative composition across dimensions to reflect the factorized Gaussian structure derived from our reformulated KAT-based Normalizing Flow. To make this multiplicative design tractable and expressive in high-dimensional spaces, we introduce a sparse hierarchical modeling strategy. Specifically, we assume that each dimension in the latent space is modeled by GMMs, where the means are parameterized via grid-based functions (analogous to KAN’s B-spline activations), and the weights are modeled independently per dimension. Furthermore, the variances are shared within a small number of high-level clusters, forming a two-layer hierarchy that significantly reduces the total number of parameters.
\begin{equation}
	p(x) = \sum_{q=1}^Q \pi_q \prod_{d=1}^D \sum_{m=1}^M w_{qdm} \frac{1}{\sqrt{2\pi} \sigma_{qd}} \exp\left\{ -\frac{(z_d - \mu_{qdm})^2}{2\sigma_{qd}^2} \right\} \left| \det \left( \frac{\partial \mathbf{z}}{\partial \mathbf{x}} \right) \right|,
\end{equation}
where $Q$ denotes the number of components in the outer mixture (also the number of covariance matrices), and $M$ denotes the number of components in the inner mixture for each dimension (also the number of means per dimension).

This construction allows us to maintain the structural clarity and interpretability of KANs, while modeling high-dimensional latent distributions.

\begin{figure}[tbp]
	\centering
	\subfigure{
		\includegraphics[width=0.22\textwidth]{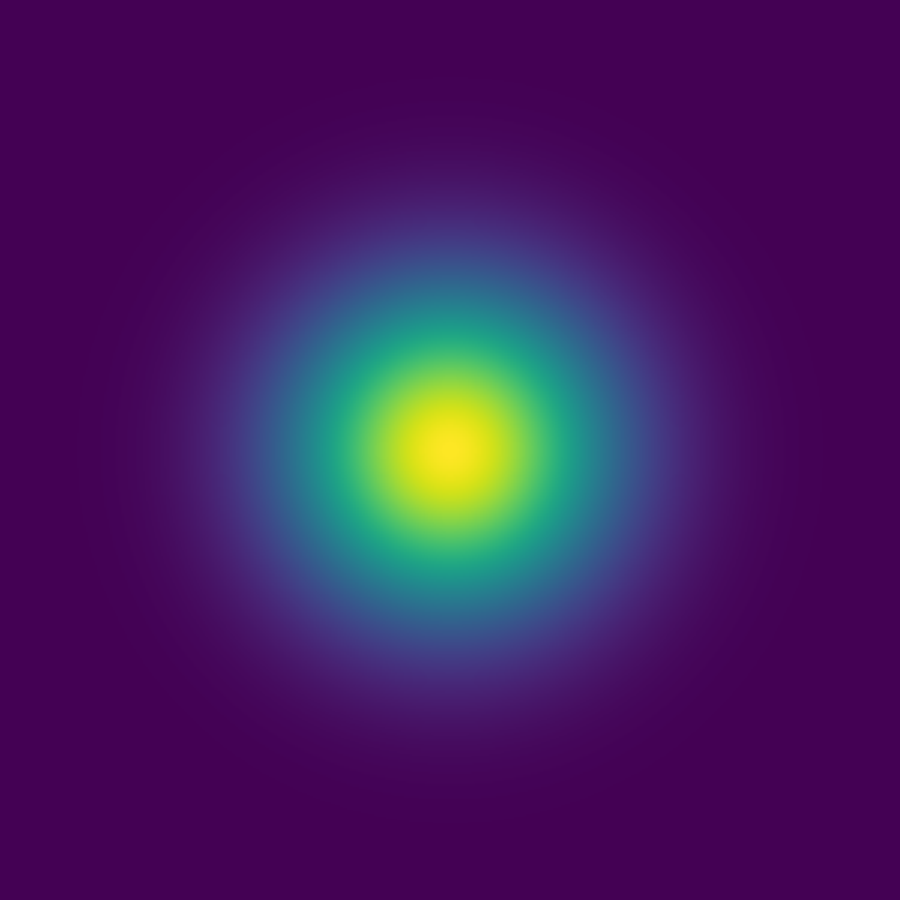}
	}
	\subfigure{
		\includegraphics[width=0.22\textwidth]{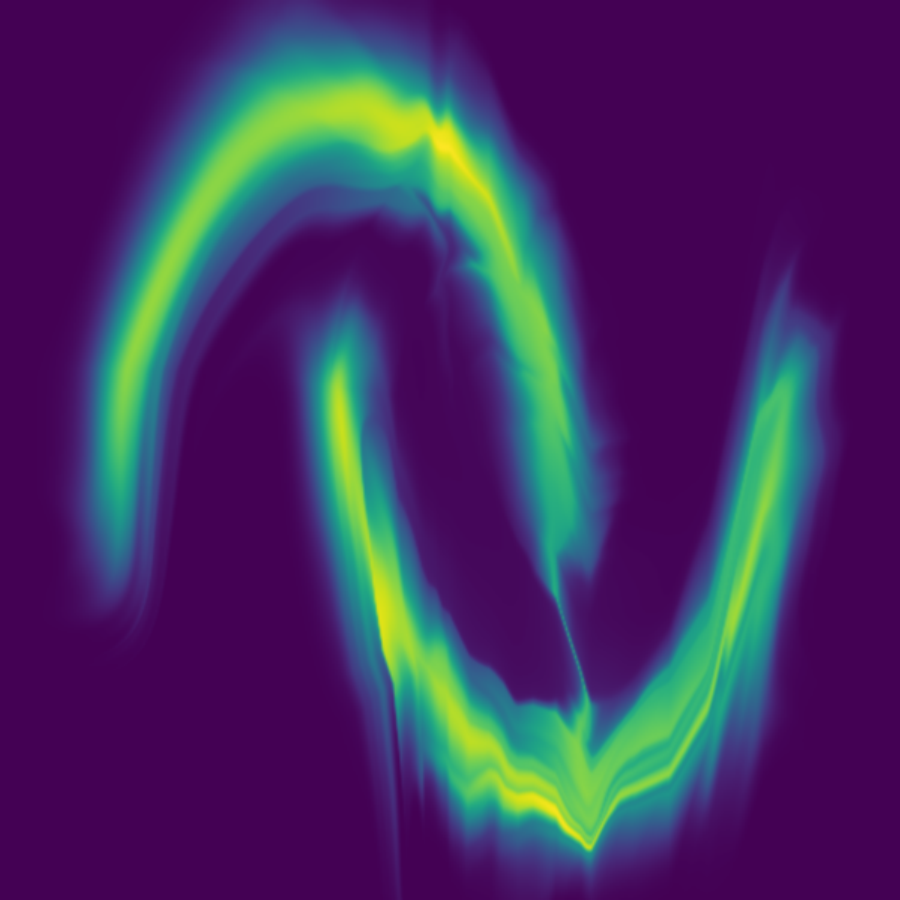}
	}
	\subfigure{
		\includegraphics[width=0.22\textwidth]{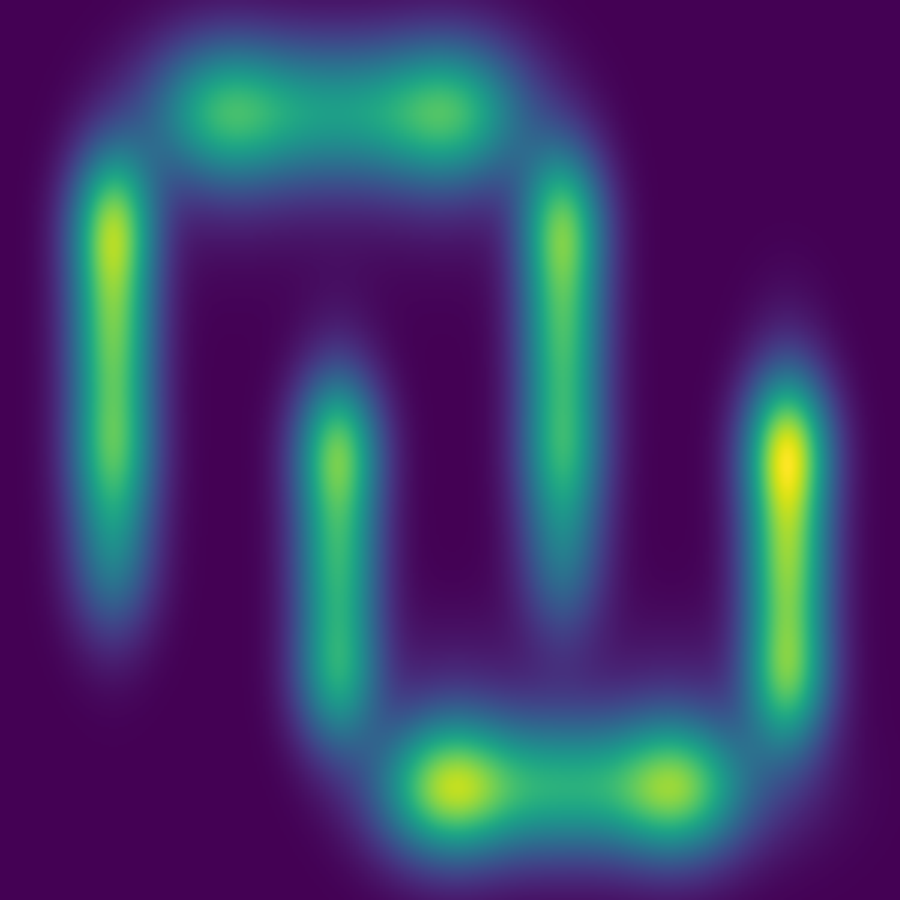}
	}
	\subfigure{
		\includegraphics[width=0.22\textwidth]{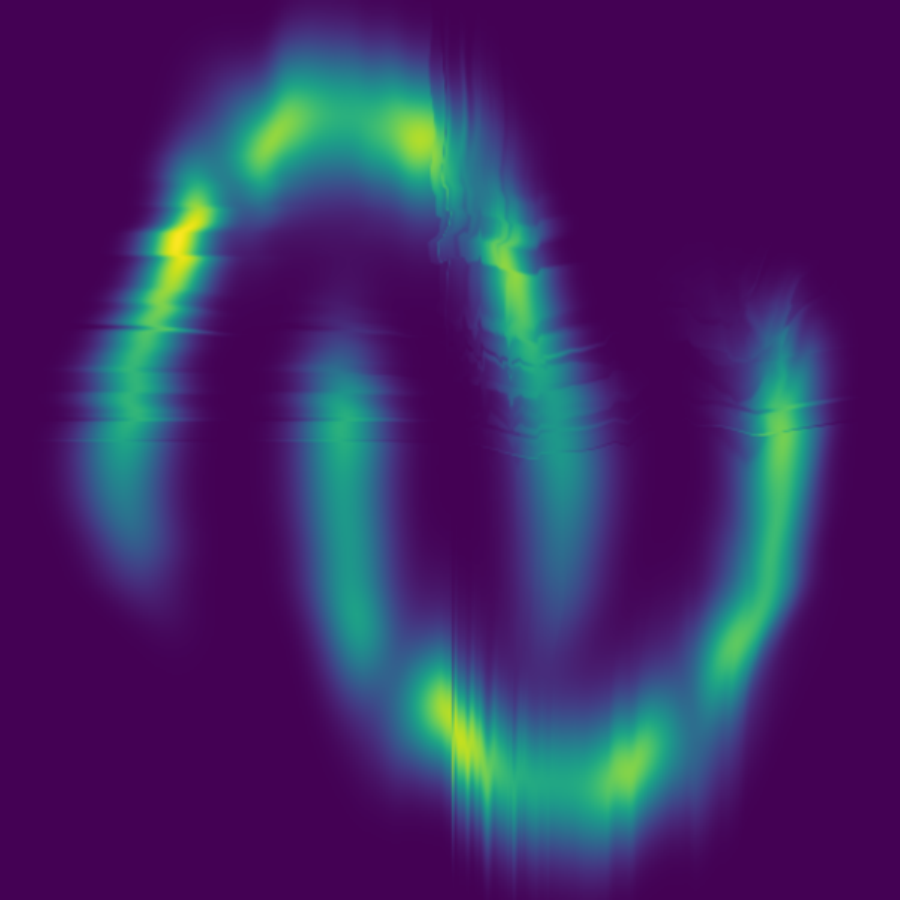}
	}
	\caption{From left to right: RealNVP-based latent space, RealNVP-based MOON’s probability, KAT-based latent space, and KAT-based MOON’s probability.}
	\label{Figure:2}
\end{figure}

\begin{figure}[tbp]
	\centering
	\includegraphics[width=\textwidth]{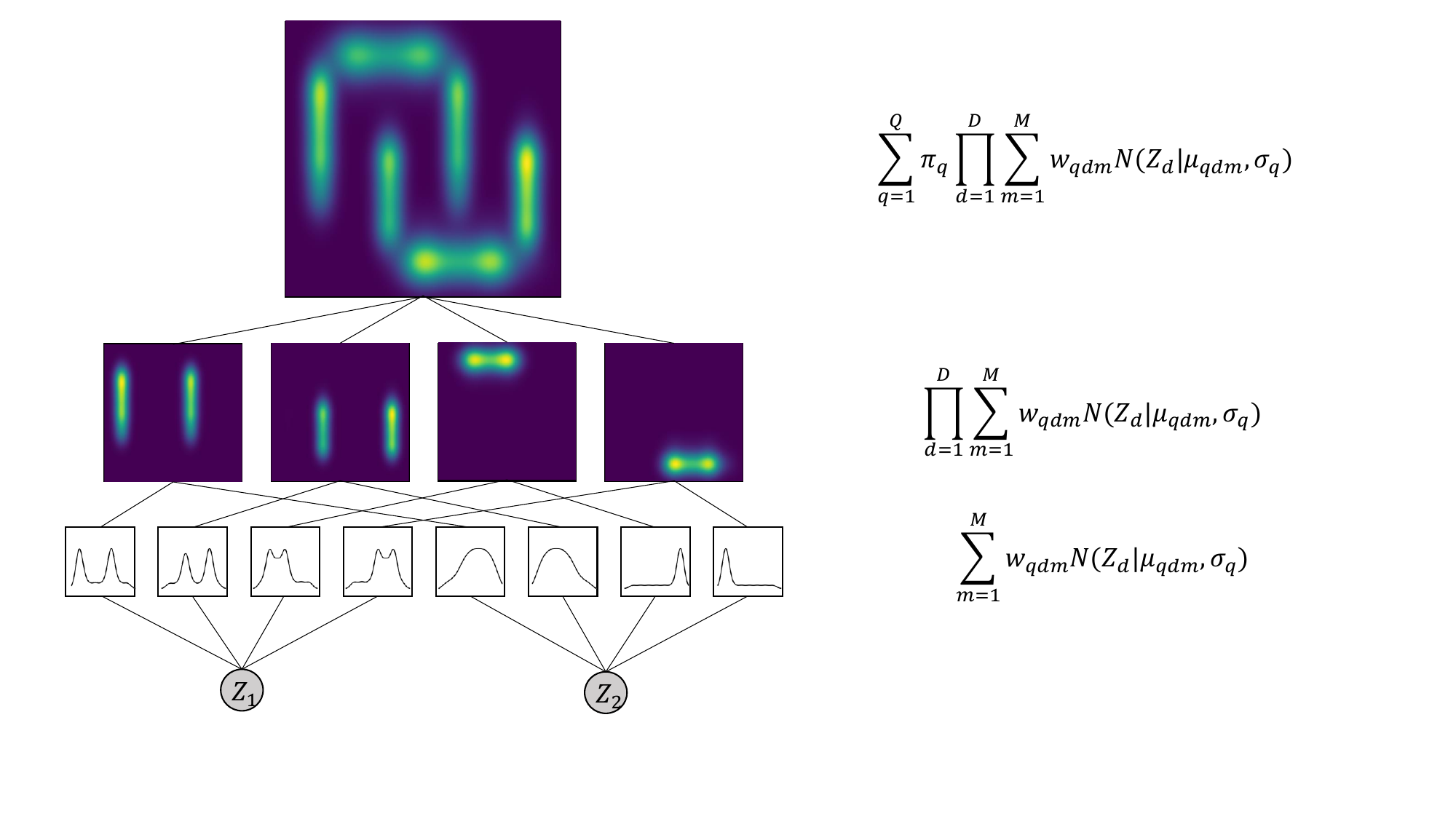}
	\caption{Neural network implementation of the structured latent space in a KAT-based normalizing flow model.}
	\label{Figure:3}
\end{figure}

\subsection{Probabilistic Modeling of the Latent Space via Latent Dirichlet Allocation}
In our model, the grid-based parameterization of the mean values across dimensions improves interpretability but inherently limits generative flexibility. When these grid-based means are treated as learnable parameters, the model often collapses to a dominant mode, leading to degeneration of the structured design and loss of diversity---a phenomenon known as mode collapse. To address this limitation, we introduce topic modeling over the discrete Gaussian functions, where the mean values are fixed and non-learnable to preserve the designed structured constraints, enabling them to be grouped into meaningful clusters. This structured organization helps preserve diversity and maintain the interpretability of the latent space. For modeling such discrete topics, LDA offers a principled probabilistic framework. By treating the mixture weights as random variables drawn from Dirichlet distributions ($Dir$), LDA enables the construction of a hierarchical Bayesian model that captures structured uncertainty and latent semantics. Importantly, it allows flexible modeling of correlations among high-dimensional discrete distributions, making it particularly well-suited for organizing latent components in a coherent and interpretable way.
\begin{equation}
	p(x) = \sum_{q=1}^Q \pi_q \prod_{d=1}^D \sum_{m=1}^M w_{qdm} \frac{1}{\sqrt{2\pi} \sigma_{qd}} \exp\left\{ -\frac{(z_d - \mu_{qdm})^2}{2\sigma_{qd}^2} \right\} \left| \det \left( \frac{\partial \mathbf{z}}{\partial \mathbf{x}} \right) \right|,
\end{equation}
where
\begin{equation}
	\bm{\pi} = (\pi_1, \pi_2, \dots, \pi_Q) \sim {Dir}(\bm{\pi} \mid \bm{\alpha}) 
	= \frac{\Gamma\left( \sum_{q=1}^Q \alpha_q \right)}{\prod_{q=1}^Q \Gamma(\alpha_q)} 
	\prod_{q=1}^Q \pi_q^{\alpha_q - 1},
\end{equation}
and similarly,
\begin{equation}
	\mathbf{w}_{qd} = (w_{qd1}, w_{qd2}, \dots, w_{qdM}) \sim {Dir}(\mathbf{w}_{qd} \mid \bm{\beta}_{qd}).
\end{equation}
$\bm{\alpha} = (\alpha_{1}, \alpha_{2}, \dots, \alpha_{Q})$ and $\bm{\beta}_{qd}$ are learnable hyperparameters of Dirichlet distributions.

\begin{figure}[tbp]
	\centering
	\includegraphics[width=\textwidth]{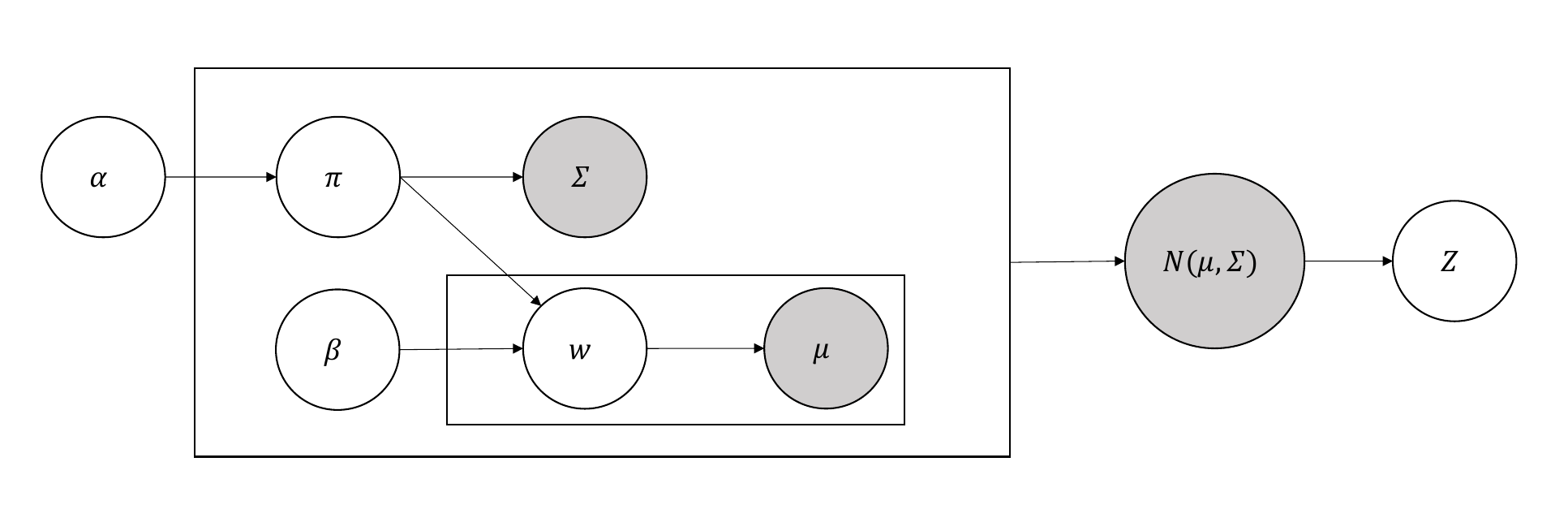}
	\caption{Statistical LDA-based modeling of the structured latent space. The latent space is constructed via a two-level LDA mechanism. In the first layer, variance parameters are assigned through topic modeling. In the second layer, mean parameters are distributed conditioned on the first layer. Finally, Gaussian samples are drawn using the obtained mean and variance parameters.}
	\label{Figure:4}
\end{figure}

In our case, the discrete weight parameters correspond to grid-based Gaussian components across dimensions. By placing Dirichlet priors over these weights, the model can flexibly control how each grid component contributes to the overall density, allowing for soft sharing across dimensions and modes. This leads to a more structured and interpretable prior, especially when the number of mixture components grows combinatorially with dimensionality. By controlling the sparsity of the weight distribution via its concentration parameters, LDA enhances the model’s ability to generalize and to capture high-order structures, especially in settings with limited supervision. Moreover, the expected log-likelihood under this mixture formulation can be computed in closed form using the $\alpha$, $\beta$ parameters, thus keeping inference efficient and tractable.

\begin{equation}
	\begin{aligned}
		&\mathbb{E}_{\bm{\pi},\bm{w}}\, p(x\mid\bm{\pi},\bm{w}) \\
		&= \sum_{q=1}^Q \frac{\alpha_q}{\sum_{q=1}^Q \alpha_{q}}
		\prod_{d=1}^D \left(
		\sum_{m=1}^M \frac{\beta_{qdm}}{\sum_{m=1}^M \beta_{qdm}}
		\frac{1}{\sqrt{2\pi}\,\sigma_{qd}}
		\exp\!\left(-\frac{(z_d-\mu_{qdm})^2}{2\sigma_{qd}^2}\right)
		\right)
		\left\lvert \det\!\left( \frac{\partial \mathbf{z}}{\partial \mathbf{x}} \right) \right\rvert.
	\end{aligned}
\end{equation}

During generation, the sampled weight vectors introduce dynamic variability and allow the model to adaptively adjust mixture components, thereby improving sample diversity and alleviating the expressiveness limitations imposed by the grid-based parameterization.

\subsection{Fractal Coupling Layers}
Normalizing flows, due to their inherently invertible transformations, require the data representation to maintain full input dimensionality throughout the flow. This structural constraint often results in parameter redundancy, especially in high-dimensional settings. However, the advantage of high-dimensional latent spaces lies in their strong information-preserving ability. Therefore, on the one hand, we aim to mitigate the computational complexity of modeling high-dimensional latent spaces; on the other hand, we want to fully leverage their capacity to retain rich information. 

Motivated by these considerations, we begin to explore novel normalizing flow architectures that balance computational efficiency and information preservation in high-dimensional latent spaces. FGMs \citep{li2025fractal} introduce a novel approach to generative modeling by recursively invoking generator modules of the same kind within itself. This recursive strategy results in a generation process that exhibits self-similar patterns across different module levels, forming a fractal-like architecture. In Generative Latent Flow (GLF) \citep{xiao2019generative}, an autoencoder is used to first compress the input into a latent representation, and a normalizing flow is then applied in the latent space to transform the distribution into a tractable form. Inspired by this approach, we are motivated to ask: Can the latent space itself be more directly and hierarchically modeled within the normalizing flow framework, forming a fractal-like structure? 

Revisiting standard NFs, we observe that each invertible transformation step can be interpreted as modeling a new latent space that is simpler and closer to the final Gaussian latent space. Rather than viewing the flow as a monolithic transformation from input to noise, we can regard each intermediate state as a distinct latent representation, culminating in the final transformation to a simple Gaussian space. This perspective opens up new possibilities for introducing structure and interpretability at every stage of the flow. After each transformation step, we can partition the latent space into several independent blocks. For each block, we employ a smaller-scale normalizing flow to model its distributions. This recursive process continues iteratively until the latent space dimensions become nearly independent.

This recursive design paradigm can be systematically incorporated into affine coupling layers to improve parameter efficiency while preserving the model’s expressive capacity. The affine coupling layer is a powerful reversible transformation where the forward function, the reverse function and the log-determinant are computationally efficient.
\begin{equation}
	\left\{
	\begin{aligned}
		y_{masked} &= x_{mask} \\
		y_{transformed} &= x_{transform} \odot \exp(s(x_{mask}) + t(x_{mask})
	\end{aligned},
	\right.
\end{equation}
where $s(\cdot)$ and $t(\cdot)$ are scale and shift functions, typically parameterized by MLPs or CNNs. The log-determinant of the Jacobian of the coupling layer operation can be computed efficiently as $\sum_j s(x_{mask})_j$. In standard affine coupling layers used in RealNVP, a checkerboard pattern is used for coupling layers before the squeezing operation while a channel-wise masking pattern is used afterward.
\begin{figure}[tbp]
	\centering
	\includegraphics[width=\textwidth]{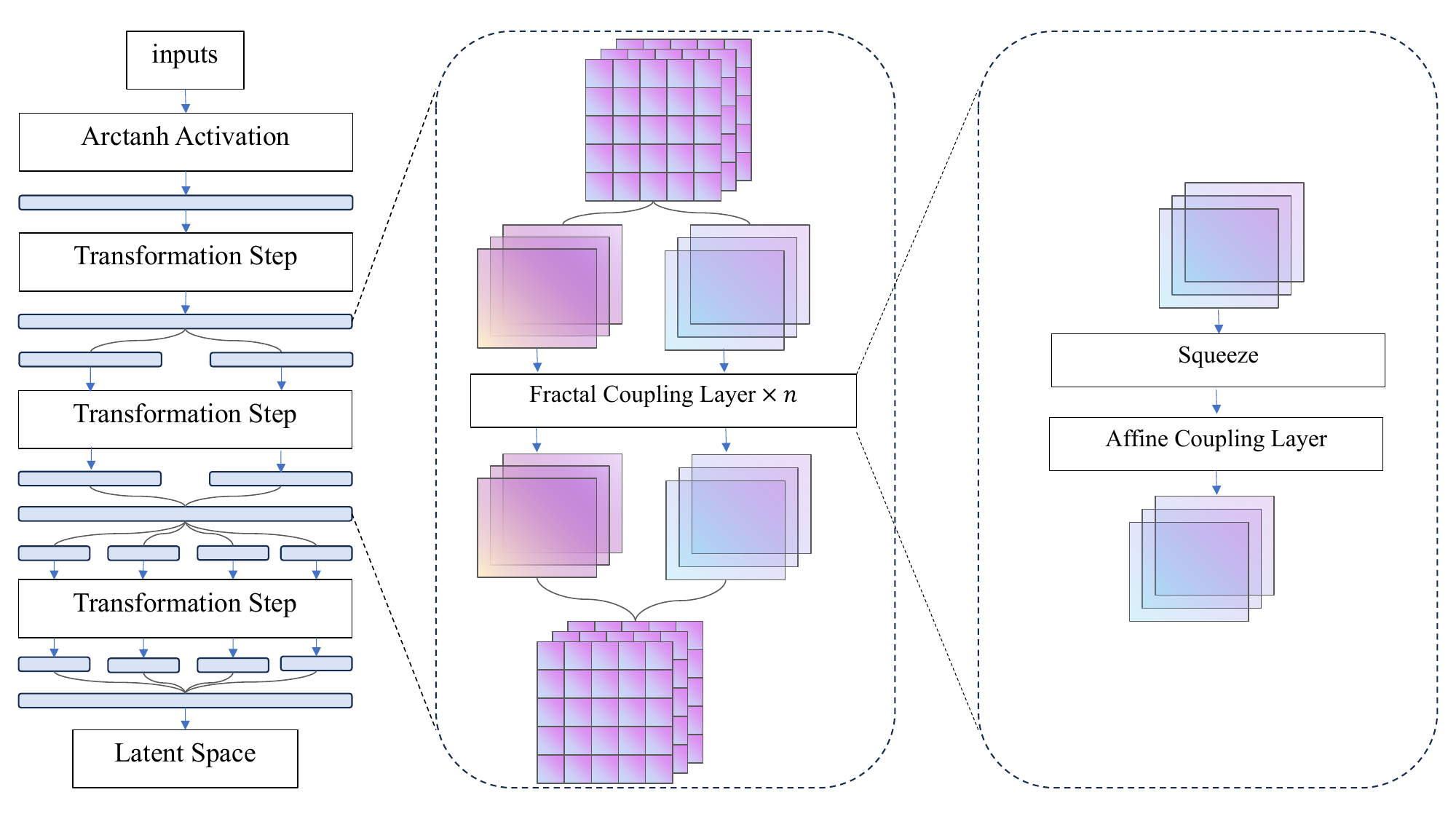}
	\caption{Network architecture of Fractal Generative Model based on affine coupling layers.}
	\label{Figure:5}
\end{figure}

This type of transformation, which only requires access to $x_{mask}$ and $x_{transform}$ and does not rely on global image context, can be naturally decomposed into a set of local transformation modules operating at different spatial resolutions in parallel. By recursively applying such modules with different receptive fields, we construct a fractal generative model that captures both fine-grained local details and global structure in a hierarchical manner. Specifically, during the transformation from real data to latent space, transformation modules with progressively decreasing spatial resolutions are used. In the generative process, the spatial resolution is gradually increased, and the global structure of the image is progressively composed through a sequence of local transformations.

To encourage interpretable and structure-preserving transformations in the flow, we adopt the same $L_2$ loss introduced in Section 2.1 to regularize the complexity of each transformation step.
\begin{equation}
	\mathcal{L}_{L_2} = \sum_i \mathbb{E}\left[ \| z_i - z_{i+1} \|_2^2 \right].
\end{equation}
\newpage
\section{Experiments}
We conduct experiments on three widely-used image datasets: MNIST, Fashion-MNIST, and the automobile class of CIFAR-10. Our evaluation focuses on three aspects: negative log-likelihood (NLL), model interpretability, and qualitative visualization results. These criteria are chosen to jointly assess the density estimation performance, structural transparency, and generative quality of our proposed method.

All images are resized to 32×32 across the three datasets. After continuous dequantization, the pixel values of images lie in $[0,256]^D$, where $D = C \times H \times W$ denotes the product of the number of channels, height, and width. In order to reduce the impact of boundary effects, we instead model the density of $arctanh\left( \frac{2\mathbf{x} - 256}{258} \right)$.

\subsection{Quantitative Experiments}
On the MNIST and FashionMNIST datasets, we adopt relatively small-scale models. The baseline model used for comparison, RealNVP, is configured with 5 affine coupling layers, each using a 3-layer MLP to parameterize the scale $s(\cdot)$ and shift $t(\cdot)$. LDANF refers to the structured latent prior introduced in Section 2.3, where one affine coupling layer in the RealNVP baseline is replaced with a single LDA layer. LDAFNF extends LDANF by incorporating the fractal coupling layers introduced in Section 2.4, where each step operates at progressively smaller spatial scales (e.g., 32×32, 16×16, etc.) to model local structures hierarchically. We further evaluate variants of the LDAFNF model with different subnetworks. All the above models are designed with controlled parameter sizes.
\begin{figure}[tbp]
	\centering
	\subfigure[MNIST]{
		\includegraphics[width=0.45\textwidth]{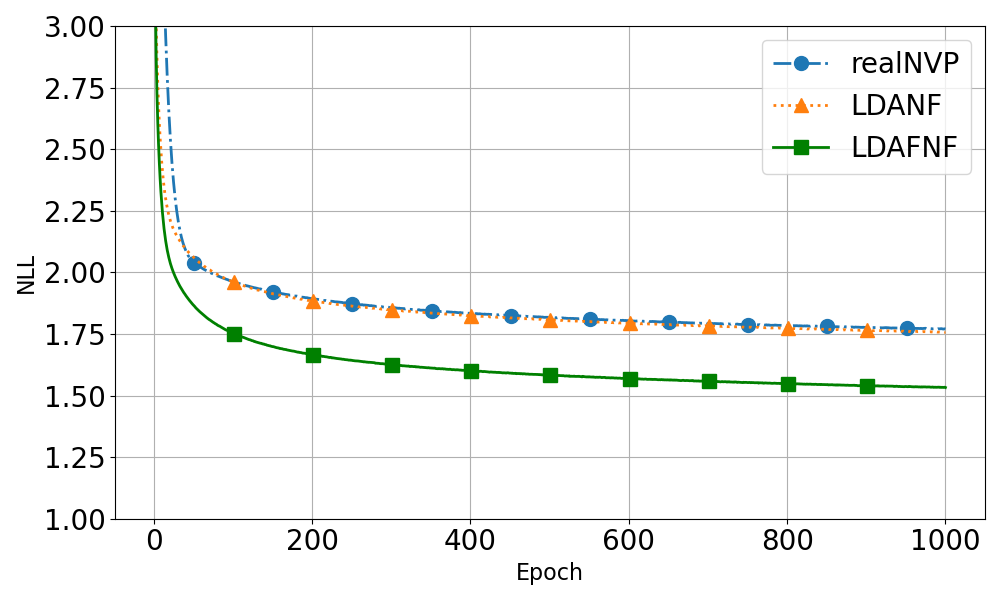}
	}
	\subfigure[MNIST]{
		\includegraphics[width=0.45\textwidth]{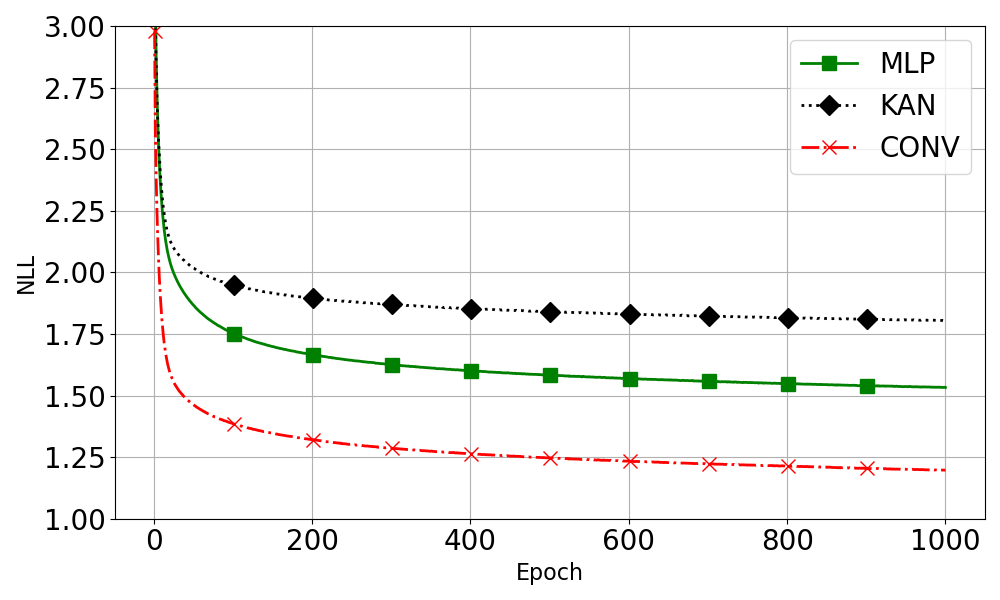}
	}
	\\
	\subfigure[FashionMNIST]{
		\includegraphics[width=0.45\textwidth]{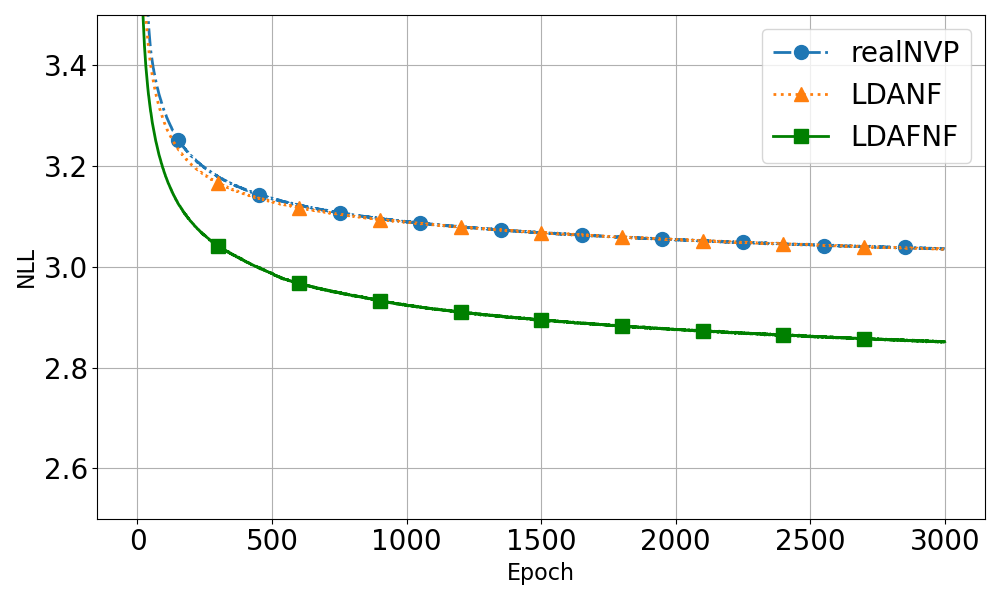}
	}
	\subfigure[FashionMNIST]{
		\includegraphics[width=0.45\textwidth]{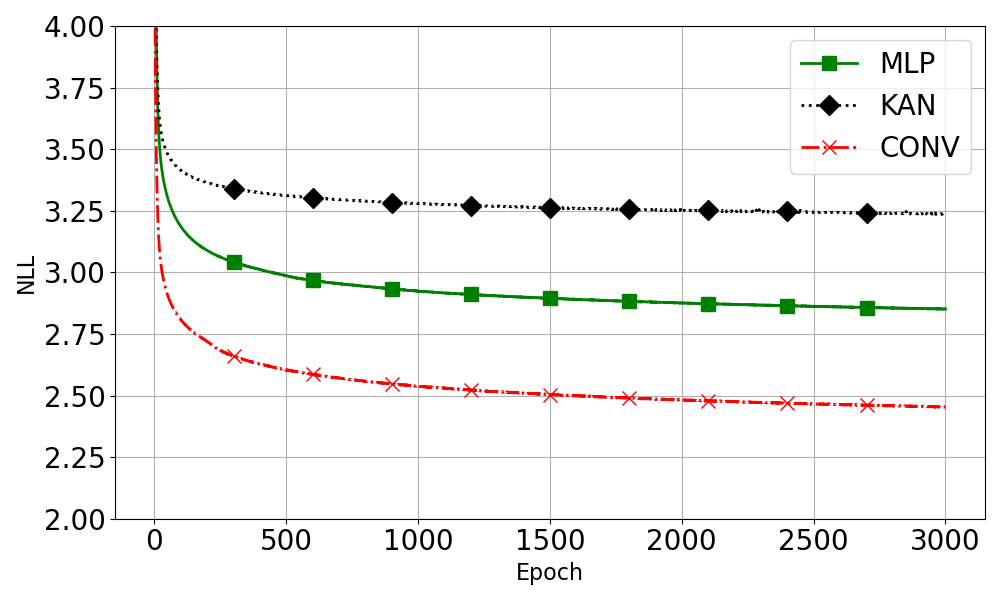}
	}
	\caption{Ablation study of the LDA layer and fractal coupling layer, along with comparisons of different subnetwork variants on MNIST and FashionMNIST.}
	\label{Figure:6}
\end{figure}

We compare the average NLL (bits per dimension) on the MNIST and FashionMNIST datasets. The results are shown in Figure 6 and Table 1. 
As can be seen, replacing one affine coupling layer with a single LDA layer does not lead to significant performance degradation in terms of NLL, highlighting the effectiveness of the proposed structured latent space representation. Meanwhile, the introduction of the fractal coupling layers achieves lower NLL and faster convergence, suggesting improved performance efficiency. Among different subnetwork configurations, convolutional subnetworks consistently achieve the highest performance efficiency.
\begin{figure}[tbp]
	\centering
	\subfigure{
		\includegraphics[width=0.45\textwidth]{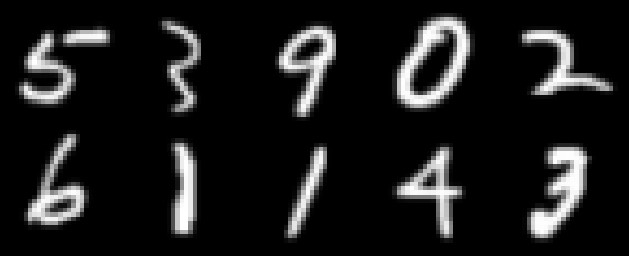}
	}
	\subfigure{
		\includegraphics[width=0.45\textwidth]{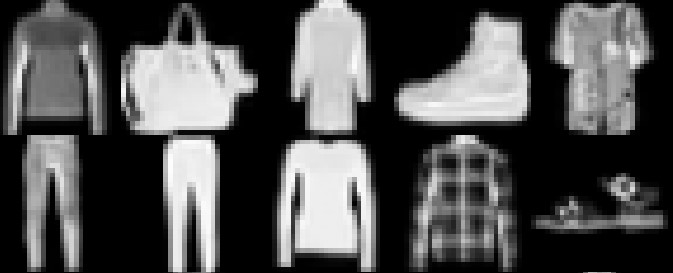}
	}
	\\
	\subfigure{
		\includegraphics[width=0.45\textwidth]{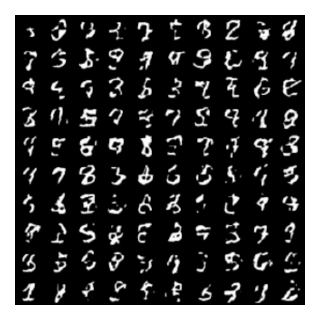}
	}
	\subfigure{
		\includegraphics[width=0.45\textwidth]{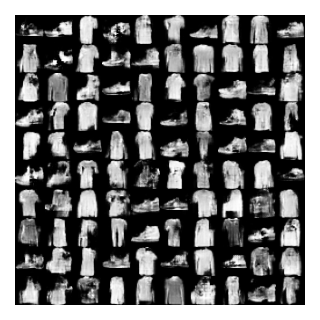}
	}
	\caption{The top row shows samples from MNIST and FashionMNIST. The bottom row shows the random samples generated from the models.}
	\label{Figure:7}
\end{figure}
\begin{table}[tbp]
	\centering
	\caption{Best results in bits per dimension of our models compared to RealNVP.}
	\label{tab:nll-results}
	\begin{tabular}{lcc}
		\toprule
		\textbf{Model} & \textbf{MNIST} & \textbf{FashionMNIST} \\
		\midrule
		RealNVP            & 1.77 & 3.04 \\
		LDANF              & 1.76 & 3.03 \\
		LDAFNF (MLP)       & 1.53 & 2.85 \\
		LDAFNF (KAN)       & 1.80 & 3.24 \\
		LDAFNF (CNN)       & \textbf{1.20} & \textbf{2.45} \\
		\bottomrule
	\end{tabular}
\end{table}

\newpage
\subsection{Qualitative Experiments}
To validate the effectiveness of our model, we conduct qualitative experiments on multiple datasets. Specifically, we present generation results on MNIST and FashionMNIST (Figure 7), demonstrate controllable generation through latent space manipulation (Figure 8), and visualize the recursive transformations in fractal coupling layers (Figure 9 and Figure 10). In addition, we evaluate the model on more complex and domain-specific data, including the automobile class from CIFAR-10 (3×32×32, Figure 11) and seismic data (1×256×64, Figure 12), to further examine its applicability. The fractal flow model successfully generates recognizable automobile images on the automobile class of CIFAR-10. For seismic data \citep{ROG2021}, leveraging the self-similar structure inherent in the signals, Fractal Flow is able to generate structurally clear signals.

\noindent\textbf{Interpretable Latent Space via LDA.} To assess the interpretability of the structured latent space introduced in Section 2.4, we visualize the topic hierarchies learned through our LDA layer. Figure 8 shows the random samples obtained from different topic modes of our model. It can be observed that these topics effectively separate the broad semantic categories of clothes and shoes, indicating that the covariance components capture high-level structure. The resulting clusters correspond to fine-grained styles of clothes and shoes, demonstrating the model's ability to disentangle low-level variations within each semantic group.

Overall, despite being trained without any labels, the model is able to uncover meaningful high-level clusters within the latent space. In particular, it successfully distinguishes shoes categories from clothing, and further distinguishes fine-grained styles within each. While some ambiguity remains—for example, trousers occasionally appear within clothing clusters—the model fails to consistently capture certain other categories in FashionMNIST. Nonetheless, the emergence of semantically coherent clusters in an unsupervised setting demonstrates the potential of our structured latent prior in learning interpretable and hierarchically organized representations.
\begin{figure}[tbp]
	\centering
	\subfigure{
		\includegraphics[width=0.45\textwidth]{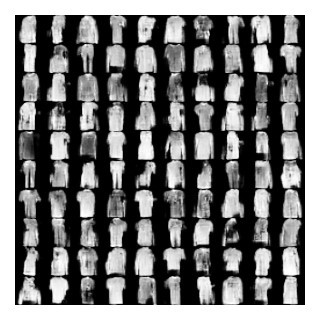}
	}
	\subfigure{
		\includegraphics[width=0.45\textwidth]{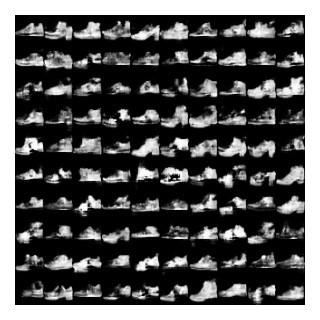}
	}\\
	\subfigure{
		\includegraphics[width=0.45\textwidth]{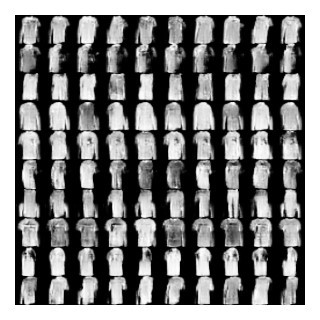}
	}
	\subfigure{
		\includegraphics[width=0.45\textwidth]{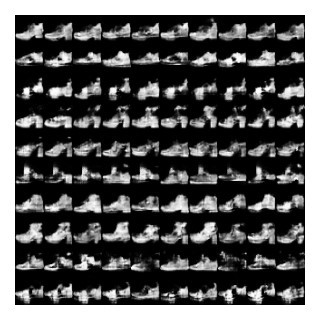}
	}	
	\caption{The top row shows random samples generated from two different covariance topics. The bottom row shows the random samples generated from different mean topics with each row corresponding to samples generated from the same mean.}
	\label{Figure:8}
\end{figure}

\newpage
\noindent\textbf{Interpretability of Fractal Coupling Layers.} To better understand how the fractal architecture contributes to the transformation process, we analyze the intermediate outputs of each fractal coupling layer. Each layer is responsible for modeling only local patterns at a specific scale, and by recursively stacking these layers, the model gradually captures more global and abstract structures.

As shown in Figure 9, we visualize the output at each transformation step using only affine coupling layers, without applying invertible 1×1 conv. This setting facilitates a clearer analysis of the contribution of each step. We observe that as the scale of the transformation modules increases, the model progressively refines the image from local details to global structure, where each step contributes to a more coherent reconstruction.
\begin{figure}[tbp]
	\centering
	\includegraphics[width=0.63\textwidth]{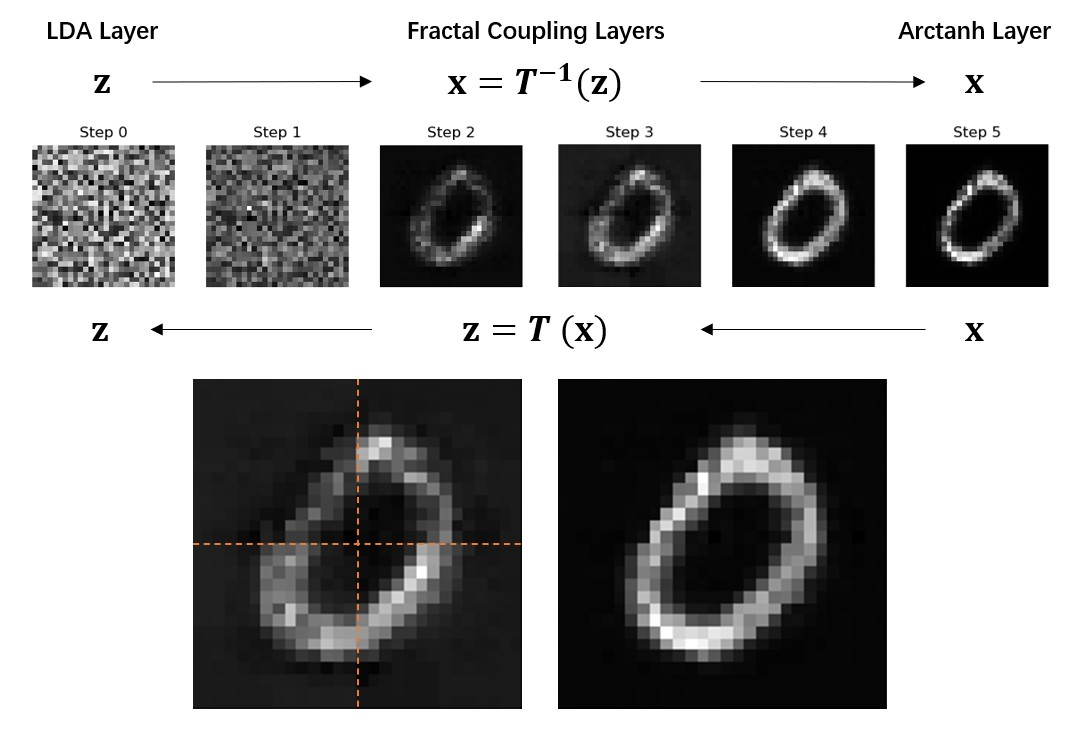}
	\caption{Visualization of the output at each transformation step on MNIST.}
	\label{Figure:9}
\end{figure}
\begin{figure}[tbp]
	\centering
	\includegraphics[width=0.6\textwidth]{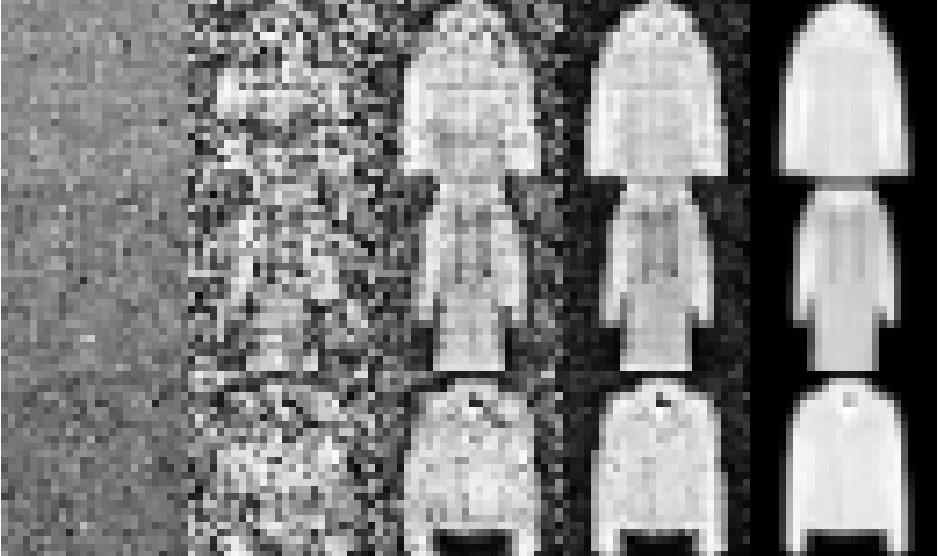}
	\includegraphics[width=0.6\textwidth]{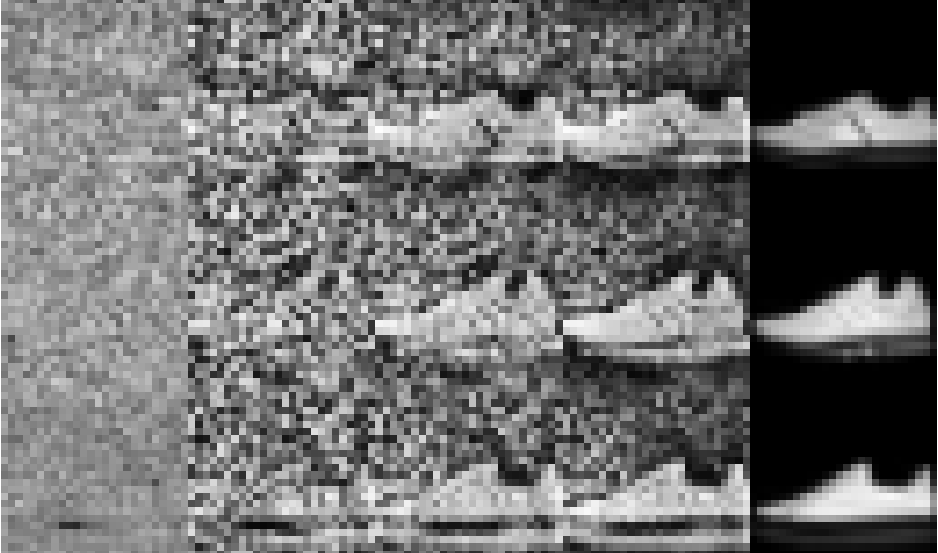}
	\caption{Visualization of the output at each transformation step on FashionMNIST.}
	\label{Figure:10}
\end{figure}
\begin{figure}[tbp]
	\centering
	\includegraphics[width=0.8\textwidth]{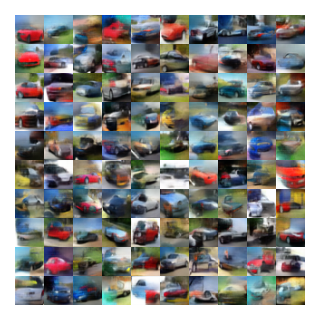}
	\caption{Random samples generated on the automobile class of CIFAR-10.}
	\label{Figure:11}
\end{figure}

\begin{figure}[tbp]
	\centering
	\subfigure[dataset]{
		\includegraphics[width=0.45\textwidth]{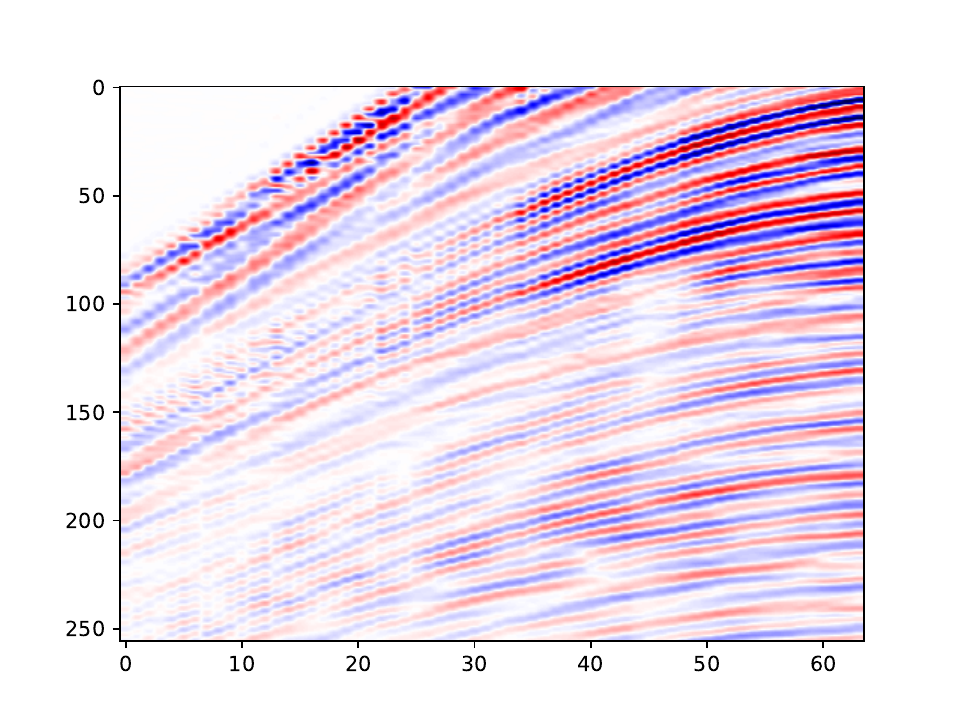}
	}
	\subfigure[dataset]{
		\includegraphics[width=0.45\textwidth]{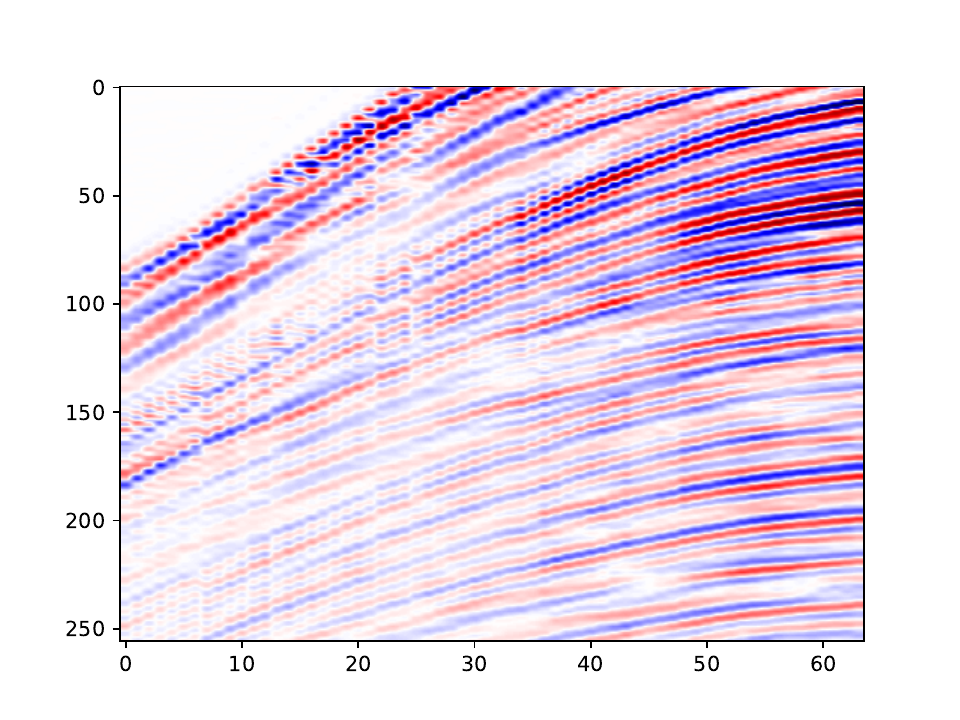}
	}
	\\
	\subfigure[temp = 0.2]{
		\includegraphics[width=0.45\textwidth]{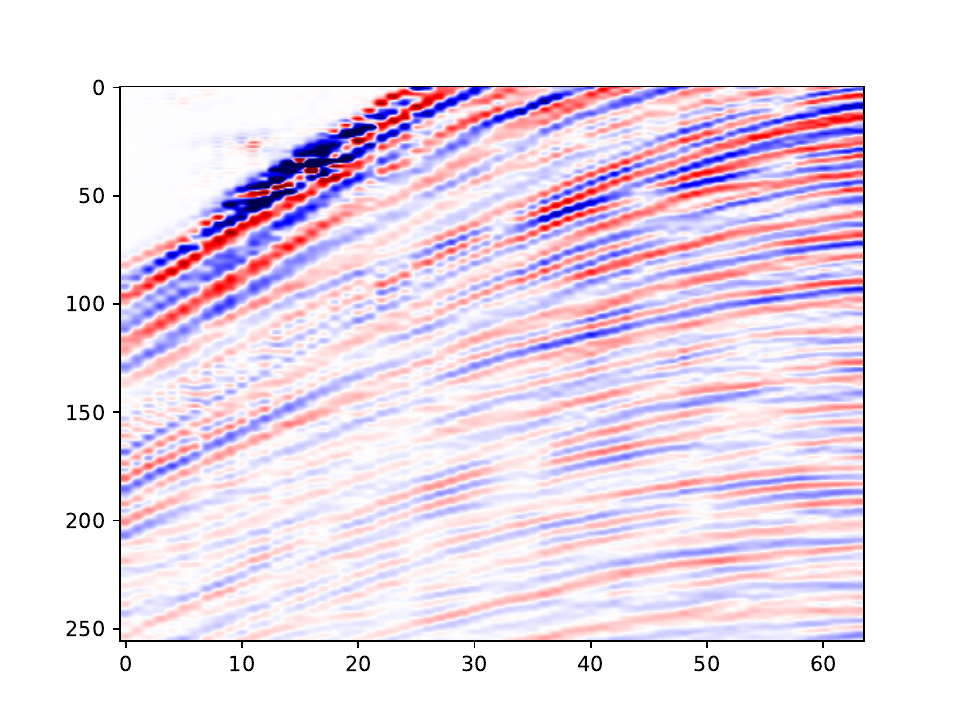}
	}
	\subfigure[temp = 0.2]{
		\includegraphics[width=0.45\textwidth]{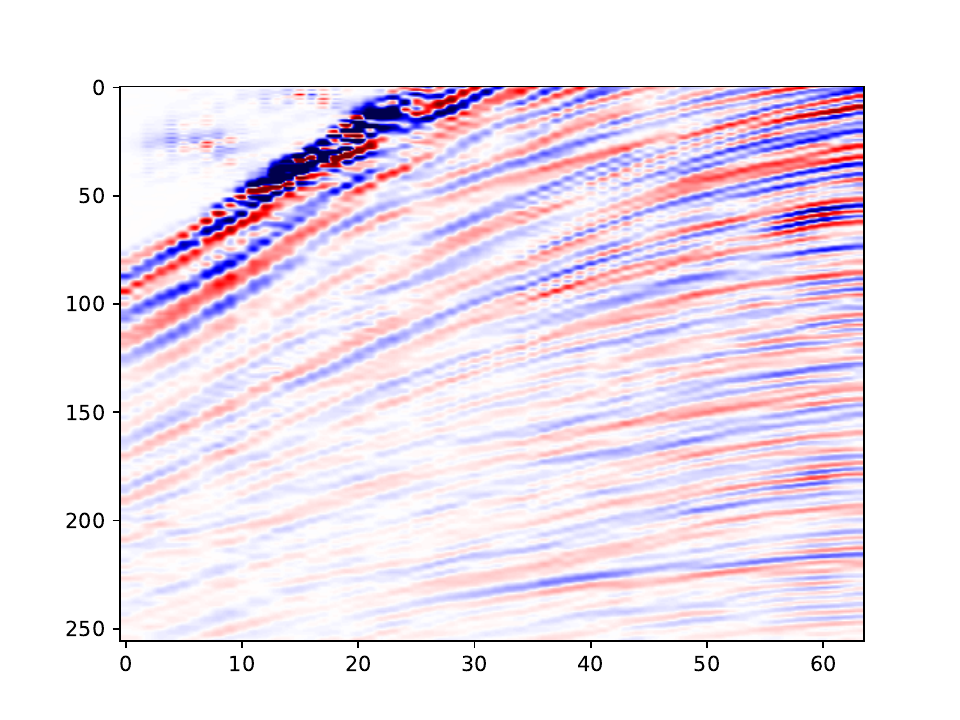}
	}
	\\
	\subfigure[temp = 0.5]{
		\includegraphics[width=0.45\textwidth]{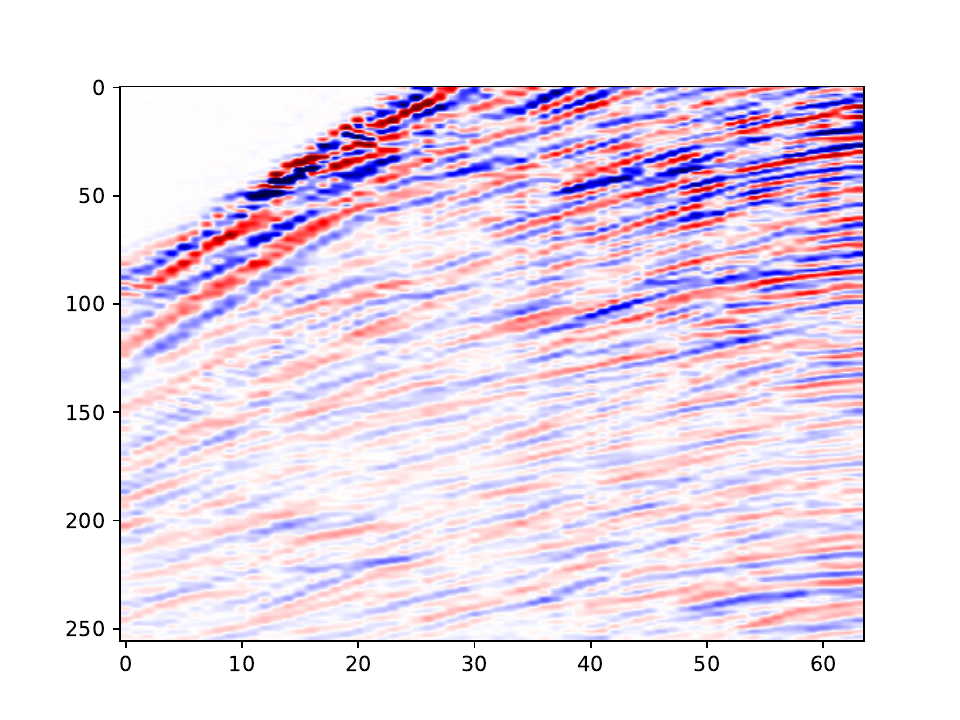}
	}
	\subfigure[temp = 0.5]{
		\includegraphics[width=0.45\textwidth]{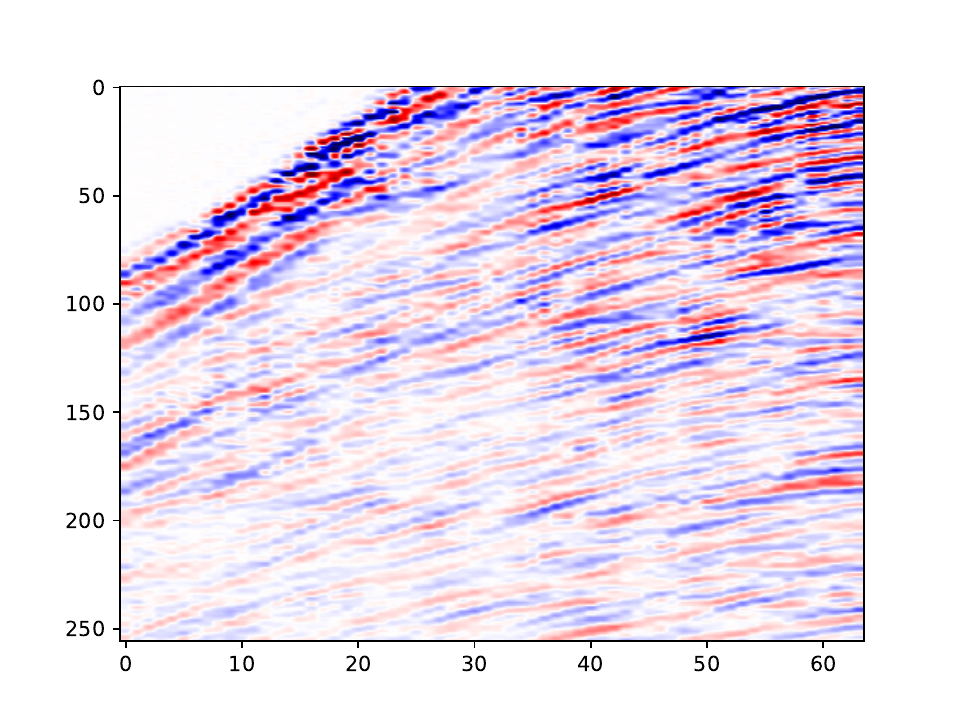}
	}
	\caption{Random samples generated on the seismic dataset.}
	\label{Figure:12}
\end{figure}

\clearpage
\section{Conclusion}
In this work, we propose a novel normalizing flow architecture (named fractal flow) that enhances both expressiveness and latent space interpretability. By integrating structured latent priors and fractal architecture, the fractal flow model demonstrates improved interpretable and controllable generalization. Experiments validate the effectiveness of our approach in both NLL evaluation and interpretability, and the model further shows strong applicability in domain-specific scenarios such as seismic data. 
Future work will be considered from several aspects: 1) Explore more adaptive coupling architectures and tighter integration between latent priors and transformation modules, 2) enhance the performance and interpretability of fractal flow via optimal transport theory, and 3) the fractal flow could be combined with diffusion models to develop a fractal flow-based diffusion framework.




\newpage








\vskip 0.2in
\nocite{*}
\bibliography{reference}

\end{document}